# Real Customization or Just Marketing: Are Customized Versions of Chat GPT Useful?

## Analysis of the Performance of a Virtual Business Statistics Professor.

**Abstract**

Large Language Models (LLMs), as the case of OpenAI ChatGPT-4 Turbo, are revolutionizing several industries, including higher education. In this context, LLMs can be personalized through a fine-tuning process to meet the student demands on every particular subject, like statistics. Recently, OpenAI has launched the possibility to fine-tune their model with a natural language web interface, enabling the possibility to create customized GPT version deliberately conditioned to meet the demands of a specific task. The objective of this research is to assess the potential of the customized GPTs that have recently been launched by OpenAI. After developing a Business Statistics Virtual Professor (BSVP), tailored for students at the Universidad Pontificia Comillas, its behavior was evaluated and compared with that of ChatGPT-4 Turbo. The results lead to several conclusions. Firstly, a substantial modification in the style of communication was observed. Following the instructions it was trained with, BSVP provided responses in a more relatable and friendly tone, even incorporating a few minor jokes. Secondly, and this is a matter of relevance, when explicitly asked for something like, "I would like to practice a programming exercise similar to those in R practice 4," BSVP was capable of providing a far superior response: having access to contextual documentation, it could fulfill the request, something beyond ChatGPT-4 Turbo's capabilities. On the downside, the response times were generally higher. Lastly, regarding overall performance, quality, depth, and alignment with the specific content of the course, no statistically significant differences were observed in the responses between BSVP and ChatGPT-4 Turbo. It appears, therefore, that customized assistants trained with prompts present advantages as virtual aids for students, yet they do not constitute a substantial improvement over ChatGPT-4 Turbo.

Eduardo C. Garrido-Merchán, Jose L. Arroyo-Barrigüete*, Francisco Borrás-Pala, Leandro Escobar-Torres, Carlos Martínez de Ibarreta, Jose María Ortíz-Lozano, Antonio Rua-Vieites

*Universidad Pontificia Comillas, Quantitative Methods Department. Madrid, Spain*

*\* Corresponding Author. jlarroyo@comillas.edu*

# Introduction

The rapid advancements in statistical generative artificial intelligence (AI) (Murphy, 2023), particularly in the realm of natural language processing and generation with the emergence of Large Language Models (LLMs) (Gozalo-Brizuela et al. 2023b, Zhao et al., 2023), based on the transformers architecture, have given birth to a new paradigm in a plethora of sectors (Gozalo-Brizuela et al. 2023a), like marketing (Fraiwan et al., 2023), higher education (Sullivan, 2023) and research (Garrido-Merchán, 2023). Among the most notable developments in this field is OpenAI's ChatGPT-4 Turbo (OpenAI, 2023), a sophisticated language model that has demonstrated remarkable capabilities in generating human-like text (Garrido-Merchán et al., 2023) and performing several tasks accurately (Peng, 2023). This technology's potential in the educational sector, especially in creating virtual teaching assistants (Baidoo-Anu et al., 2023) is immense. However, the effectiveness and practical utility of these AI models, when fine-tuned and customized for specific educational purposes, remain areas of burgeoning research.

The concept of customized generative artificial intelligence, particularly in LLMs like ChatGPT-4, involves fine-tuning the model on specific multimodal data, datasets or with tailored prompts to better suit particular tasks. Intuitively, the LLM capacity exceeds the complexity of a particular task, such as being a virtual instructor, so by fine-tuning the model, we condition its behavior to make him more suitable to only provide solutions to that particular task. This process is hypothesized to make these models more effective in specialized tasks, such as functioning as a virtual professor in a specific academic discipline. The recent introduction of a natural language web interface for fine-tuning by OpenAI has made this process more accessible, democratizing its adoption in diverse fields.

The relevance of this research is motivated by the growing demand for personalized learning experiences in higher education and democratizing it, being reachable to any specific group of persons in need. Customized AI models promise to provide more relatable, engaging, and personalized learning interactions, potentially transforming the traditional educational landscape. However, as this experience is only provided by OpenAI's ChatGPT, the extent to which these customized models genuinely enhance the learning experience and improve educational outcomes, as opposed to being mere marketing claims, warrants rigorous investigation.

This study, therefore, focuses on evaluating the efficacy of a customized GPT version of ChatGPT-4 Turbo, developed as a Business Statistics Virtual Professor (BSVP), specifically for statistics students at the Business Faculty of Universidad Pontificia Comillas. By comparing the performance of this tailored model with the standard ChatGPT-4 Turbo in this particular task, this research aims to provide insights into the actual benefits and limitations of AI customization in an educational context. Such an analysis is crucial in understanding whether these AI-driven educational tools represent a significant advancement or if they are merely incremental improvements over existing technology.

The organization of this paper is the following one. First, we begin with a related work section where we briefly describe the use of LLMs in the higher education sector. Then, we continue with a material and methods section, where we describe the prompt and materials used to create the customized BSVP GPT and we describe the methodology that we have followed to determine the effectiveness of BSVP GPT with respect to ChatGPT-4 Turbo. We continue this work with a technical section where we briefly describe the GPT architecture. Then, we include a results and discussion section where we describe the results that we have obtained to evaluate the performance of BSVP GPT for the statistics class. Finally, we finalize this paper with a conclusions and further work section where we summarize our findings and propose new research directions.

## Related Work

The integration, challenges and opportunities of Generative AI into higher education, especially in the context of teaching, has garnered considerable attention in recent years (Michel-Villarreal et al., 2023). This section reviews the latest research in the field (Lo et al., 2023), emphasizing studies that explore the role of generative AI in teaching, its application as a virtual assistant, and its contribution to academic research.

Recent studies in this domain have focused on the efficacy of generative AI in enhancing teaching methodologies (Baidoo-Anu et al., 2023). These works highlight the potential of AI in personalizing learning experiences, providing real-time feedback, and augmenting traditional teaching practices (Kasneci et al., 2023, Zhai et al., 2022). For example, ChatGPT has been proven useful for lifelong learning (Rawas, 2023), as, for instance, it can readapt the teaching lessons to the latest advances of rapidly changing technologies.

However, generative AI has also raised a debate about evaluation methodologies of higher education (Anders et al., 2023), as its content generation can be used by students to cheat easily (Cotton et al., 2023). For example, evaluations done by professors have changed to adapt to this paradigm shift as, for instance, traditional assessments are easier to cheat as ever with generated content of Generative AI (Rudolph, 2023).

Another significant area of research involves the use of generative AI as virtual assistants in educational settings (Chheang et al., 2023). These studies explore the capabilities of AI assistants in managing student inquiries, offering personalized tutoring, and facilitating learning outside the traditional classroom environment (Ruiz-Rojas et al., 2023).

Finally, the role of generative AI in academic research (Xames et al., 2023) has been an area of growing interest (Rahman et al., 2023). These investigations delve into how AI can assist in data analysis, brainstorming of ideas, literature review, synthetic data generation, text simplification and even in assisting to write some sections of research papers, thereby augmenting the research capabilities of scholars and students alike (Garrido-Merchán, 2023).

## Generative Pretrained Transformers (GPTs)

The evolution of Generative Pretrained Transformers (GPTs) (Radford et al., 2018) has produced a paradigm shift in the democratization of natural language processing (NLP) (Chowdhary et al., 2020). The journey began with the original GPT model (Radford et al., 2018), introduced by OpenAI, whose novelty includes unsupervised learning to predict the next word in a sentence, not only supervised learning as was done before. More concretely, GPT's methodology encompassed a dual-phase process: an initial "pre-training" stage using an unsupervised generative approach to establish baseline parameters through language modeling, followed by a "fine-tuning" stage, where these parameters were refined and tailored to a specific task in a supervised, discriminative manner.

This model laid the groundwork for more advanced iterations. GPT-2 (Radford et al., 2019), marked a significant leap with its 1.5 billion parameters and more engineering tricks, demonstrating enhanced text generation capabilities and enabling the hypothesis that scale was all that natural language processing needs. However, its behavior showed clues of underfitting, being its capacity, despite its 1.5 billion parameters, too simple for the complexity of the corpus. Motivated by this underfitting hypothesis, OpenAI launched GPT-3 (Brown et al., 2020), revolutionizing the field with its 175 billion parameters and offering unprecedented language understanding and generation proficiency. It is important to emphasize that each iteration of GPT has built upon the transformer architecture (Vaswani et al., 2017). This architecture abandoned the recurrent layers used in previous models, relying instead on a self-attention mechanism that allowed the model to weigh the significance of different parts of the input data.

ChatGPT then emerged as a GPT 3.5 version that optimized the conversational experience with a user, being ChatGPT-4 (OpenAI, 2023) and ChatGPT-4 Turbo the latest iteration in this series, standing out with its enhanced capabilities and efficiency, in comparison with GPT-3 (Peng et al., 2023). This version maintains the core transformer architecture but introduces several optimizations for speed and performance. ChatGPT-4 Turbo is designed to handle more complex queries with greater accuracy and speed, making it particularly suitable for real-time applications. The model's ability to understand and generate human-like text is underpinned by its advanced training algorithms and vast corpora, which encompass a wide array of human knowledge and language nuances.

A critical component in the development of GPT models, especially ChatGPT-4 Turbo, that explains its outstanding behavior, is Reinforcement Learning from Human Feedback (RLHF) (Christiano et al., 2017). This training approach involves fine-tuning models based on feedback from human trainers. Initially, the model generates responses based on its pretraining; these responses are then evaluated by humans who provide ratings or improved versions of the responses. The model is subsequently retrained to prefer the human-approved responses. This method ensures that the model's outputs align more closely with human preferences, leading to more accurate and contextually appropriate responses, that now with the fine-tune versions of ChatGPT like BSVP can gain even more importance.

Precisely, the fine-tuning process in GPT models allows for the customization of the base model to suit specific applications or domains. The fine-tuning process involves training the pre-existing model on a smaller, domain-specific dataset, enabling the model to adapt its responses to the nuances of a particular field or user requirement. Fine-tuning can significantly enhance the model's performance in specialized tasks by adjusting its outputs to be more aligned with the specific content, style, or tone required by the application. This customization is pivotal for applications like educational tools, where the model needs to understand and respond appropriately to subject-specific queries and pedagogical requirements.

## Material and Methods

Initially, a virtual assistant for Statistics courses taught at Universidad Pontificia Comillas was created. The assistant was instructed via prompt with specific directions regarding communication style. Additionally, contextual documentation was provided: two books written by three professors of the subject and signatories of this research (Borrás-Pala et al., 2019a, 2019b), as well as the R programming practices document, prepared by another three different professors, who are also authors of this work. Over three days, the system was tested by two of the authors, progressively refining the prompt until they achieved a version they considered acceptable. At that point, the evaluation began, which is described below.

The study was conducted through the assessment of BSVP's response quality by the five professors who signed this work but did not participate in the generation and subsequent adjustment of the prompt. Specifically, the work was carried out in four different stages. Firstly, each professor collected between 15 and 30 questions posed by students of the "Statistics and Probability" and "Business Statistics" courses, which are taught across seven different degrees. In most cases, these were second-year courses, and in some instances, third-year courses. All questions had to be real inquiries made by students during classes or tutoring sessions. This is a highly relevant aspect, as students often struggle to clearly and specifically articulate their doubts (e.g., "I don't understand what this Student's t is about"; "In the Poisson binomial, how is lambda calculated?"): it's important to evaluate the system's ability to competently respond to these kinds of questions. If BSVP is to act as a virtual assistant for students, it should be able to answer such questions despite their ambiguity, lack of definition or even errors in the question itself. In the second stage, each question was posed to ChatGPT-4 Turbo and BSVP, noting down both complete responses. To ensure comparability, there were no follow-up questions or clarifications; the first provided response was copied, whether satisfactory or not. In the third phase, the professors who had not participated in the generation and adjustment of the prompt, evaluated the responses from ChatGPT-4 Turbo and BSVP, scoring them on a scale of 0 to 10. The choice of this specific scale responds to the characteristics of the Spanish university system, where it is the default scale used to evaluate university students. Therefore, the professors responsible for this evaluation are very familiar with this scale. It is important to note that the evaluation was blind, as each professor assessed both responses without knowing who the author was (ChatGPT-4 Turbo or BSVP). Only the two professors who did not participate in the evaluation had this information. Specifically, three different dimensions were evaluated: quality of the response (clarity, conciseness, etc.); depth of the response (to what extent it is as complete as possible); and personalization (degree of closeness to the way the subject is taught at the university where the study was conducted). Finally, in the fourth stage, a statistical comparison of the results obtained by both systems was carried out.

## Results and discussion

Starting with a qualitative assessment, a substantial modification in the style of communication was indeed observed. As per its training, BSVP provided responses in a much more approachable and friendly tone. In fact, it often began responses with phrases like "Dear ICADE[1] student," "This question you ask is very interesting," or "Excellent question, my dear ICADE student!" The farewells were also more cordial ("a big hug," "I hope this has helped you") and occasionally, it incorporated small jokes ("Perhaps your ICADE teacher might say something different, though I doubt it. But after all, they are human, and I am not, so I know much more than them ;-)")[2]. A greater conciseness in the responses was also generally observed, which was instructed in the training prompt.

---

[1] ICADE - Instituto Católico de Administración y Dirección de Empresas (Catholic Institute of Business Administration and Management). It is the name of the business school of the Universidad Pontificia Comillas, where the study was conducted.
[2] To ensure that the evaluation was blind, all these phrases were removed from the responses, so that the evaluators were not aware of them

A highly relevant aspect is that, when explicitly asked for something like "I would like to practice a programming exercise similar to those in R programming practice 3," BSVP was capable of providing a much superior response: having access to contextual documentation, it was able to address the request, something that was not possible for ChatGPT-4 Turbo[3]. However, as a trade-off, the response times were generally longer.

Regarding the content, a total of 136 questions were obtained, which, as mentioned, were evaluated according to three different dimensions: quality, depth, and personalization. Figure 1 shows the corresponding bar plots.

**Figure 1:** Bar plot of the scores obtained by BSVP and ChatGPT-4 Turbo in each of the three dimensions analyzed

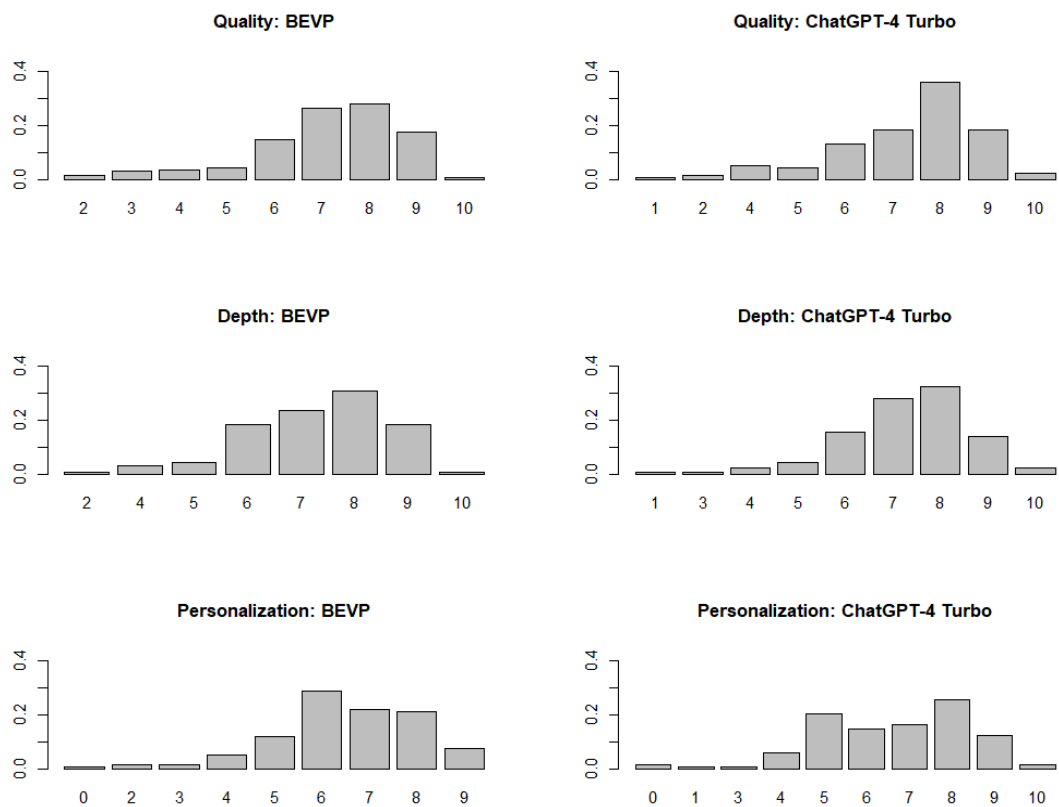

The comparative analysis of the performance of both systems (see Table 1) suggests that there are no significant differences in any dimension. The most interesting aspect is the absence of differences in personalization (degree of closeness to the way the subject is taught at the university where the study was conducted), indicating that the contextual documentation has not really served to offer adapted content. As already mentioned, this documentation is very useful when the question includes an explicit reference to course content (i.e., "I would like to practice a programming exercise similar to those in R programming practice 3"), as it allows BSVP to respond competently. However, in more general questions like those included in this evaluation, which do not require the consultation of contextual documentation, there are no differences between BSVP and ChatGPT-4 Turbo.

---
[3] Logically, questions of this nature were not included in the evaluation.

**Table 1:** Results obtained in each dimension. Mean, standard deviation (sd), and t-test for mean difference.

|  | Quality | Depth | Personalization |
|---|---|---|---|
| **BSVP: mean (sd)** | 7.12 (1.60) | 7.30 (1.36) | 6.50 (1.57) |
| **ChatGPT-4 Turbo: mean (sd)** | 7.30 (1.62) | 7.29 (1.39) | 6.64 (1.84) |
| **t-test** | t = -1.098<br>df = 135<br>p-value = 0.274 | t = 0.096<br>df = 135<br>p-value = 0.924 | t = -0.855<br>df = 135<br>p-value = 0.394 |

## Conclusions

The main conclusions of this research can be summarized in three key ideas. Firstly, differences in communication style are indeed noticeable. Training via prompt has created a virtual assistant whose style is clearly distinct from that of ChatGPT-4 Turbo. Secondly, BSVP has a significant advantage over ChatGPT-4 Turbo: its contextual documentation allows it to respond to specific queries about course content, something ChatGPT-4 Turbo cannot do. This is not a minor aspect, as students often pose questions in this manner (e.g., "could you provide an example of a problem like those in chapter 4"; "I don't understand the first part of the R programming practice 6"). Lastly, regarding general content, no significant differences are evident. That is, ChatGPT-4 Turbo is capable of answering any query in a manner similar to BSVP. However, we must consider that we are dealing with a subject that is quite basic and for which there is an enormous amount of information. Therefore, the responses cannot vary much in terms of quality and depth.

It seems, therefore, that customization via prompt does represent a certain improvement, relevant if students value a more affable communication style and the possibility of asking about specific topics from the course content. For a student looking in their virtual assistant for a system to help resolve his/her doubts, BSVP offers no advantage over ChatGPT-4 Turbo.

The main limitation of this study is that it is very preliminary. To verify the conclusions reached, it would be necessary to conduct an experiment in which the students themselves, as the end-users of BSVP, would evaluate the responses provided by both systems. However, such an experiment is not without difficulties, as students often may not be in a position to accurately assess the responses. For instance, it is quite possible that they would favor a brief and direct answer over a longer and more complex one, even though the latter might be correct and the former not. They probably would not be able to discriminate based on the veracity of the result. Additionally, there is a high likelihood that students would be influenced by the style of communication, so their assessment of the responses might be affected by the way each system responds. Nevertheless, we believe these and other challenges can be overcome with proper experimental design, which is why we propose as a future line of research to delve deeper into the differences between both systems considering the students' perspective. Another possible line of research is to assess the behavior in more advanced and therefore more specialized subjects, belonging to higher courses of the degree or even to postgraduate subjects.

## References

Anders, B. A. (2023). Is using ChatGPT cheating, plagiarism, both, neither, or forward thinking?. *Patterns*, *4*(3).

Baidoo-Anu, D., & Ansah, L. O. (2023). Education in the era of generative artificial intelligence (AI): Understanding the potential benefits of ChatGPT in promoting teaching and learning. *Journal of AI*, *7*(1), 52-62.

Borrás-Pala, F., Martinez de Ibarreta, C., Escobar-Torres, L. (2019a). *Estadística Empresarial en 101 ejemplos (volumen I)* EV Services.

Borrás-Pala, F., Martinez de Ibarreta, C., Escobar-Torres, L. (2019b). *Estadística Empresarial en 101 ejemplos (volumen II)* EV Services.

Brown, T., Mann, B., Ryder, N., Subbiah, M., Kaplan, J. D., Dhariwal, P., ... & Amodei, D. (2020). Language models are few-shot learners. *Advances in neural information processing systems*, *33*, 1877-1901.

Chheang, V., Marquez-Hernandez, R., Patel, M., Rajasekaran, D., Sharmin, S., Caulfield, G., ... & Barmaki, R. L. (2023). Towards anatomy education with generative AI-based virtual assistants in immersive virtual reality environments. *arXiv preprint arXiv:2306.17278*.

Chowdhary, K., & Chowdhary, K. R. (2020). Natural language processing. *Fundamentals of artificial intelligence*, 603-649.

Christiano, P. F., Leike, J., Brown, T., Martic, M., Legg, S., & Amodei, D. (2017). Deep reinforcement learning from human preferences. *Advances in neural information processing systems*, *30*.

Cotton, D. R., Cotton, P. A., & Shipway, J. R. (2023). Chatting and cheating: Ensuring academic integrity in the era of ChatGPT. *Innovations in Education and Teaching International*, 1-12.

Fraiwan, M., & Khasawneh, N. (2023). A Review of ChatGPT Applications in Education, Marketing, Software Engineering, and Healthcare: Benefits, Drawbacks, and Research Directions. *arXiv preprint arXiv:2305.00237*.

Garrido-Merchán (2023). Best uses of ChatGPT and Generative AI for computer science research. *arXiv preprint arXiv:*2311.11175

Garrido-Merchán, E. C., Arroyo-Barrigüete, J. L., & Gozalo-Brihuela, R. (2023). Simulating HP Lovecraft horror literature with the ChatGPT large language model. *arXiv preprint arXiv:2305.03429*.

Gozalo-Brizuela, R., & Garrido-Merchán, E. C. (2023a). A survey of Generative AI Applications. *arXiv preprint arXiv:2306.02781*.

Gozalo-Brizuela, R., & Garrido-Merchan, E. C. (2023b). ChatGPT is not all you need. A State of the Art Review of large Generative AI models. *arXiv preprint arXiv:2301.04655*.

Kasneci, E., Seßler, K., Küchemann, S., Bannert, M., Dementieva, D., Fischer, F., ... & Kasneci, G. (2023). ChatGPT for good? On opportunities and challenges of large language models for education. *Learning and individual differences*, *103*, 102274.

Lo, C. K. (2023). What is the impact of ChatGPT on education? A rapid review of the literature. *Education Sciences*, *13*(4), 410.

Michel-Villarreal, R., Vilalta-Perdomo, E., Salinas-Navarro, D. E., Thierry-Aguilera, R., & Gerardou, F. S. (2023). Challenges and Opportunities of Generative AI for Higher Education as Explained by ChatGPT. *Education Sciences*, *13*(9), 856.

Murphy, K. P. (2023). *Probabilistic machine learning: Advanced topics*. MIT press.

OpenAI, 2023. GPT-4 technical report, [https://cdn.openai.com/papers/gpt-4.pdf](https://cdn.openai.com/papers/gpt-4.pdf)

Peng, B., Li, C., He, P., Galley, M., & Gao, J. (2023). Instruction tuning with gpt-4. *arXiv preprint arXiv:2304.03277*.

Radford, A., Narasimhan, K., Salimans, T., & Sutskever, I. (2018). Improving language understanding by generative pre-training.

Radford, A., Wu, J., Child, R., Luan, D., Amodei, D., & Sutskever, I. (2019). Language models are unsupervised multitask learners. *OpenAI blog*, *1*(8), 9.

Rahman, M. M., & Watanobe, Y. (2023). ChatGPT for education and research: Opportunities, threats, and strategies. *Applied Sciences*, *13*(9), 5783.

Rawas, S. (2023). ChatGPT: Empowering lifelong learning in the digital age of higher education. *Education and Information Technologies*, 1-14.

Rudolph, J., Tan, S., & Tan, S. (2023). ChatGPT: Bullshit spewer or the end of traditional assessments in higher education?. *Journal of Applied Learning and Teaching*, *6*(1).

Ruiz-Rojas, L. I., Acosta-Vargas, P., De-Moreta-Llovet, J., & Gonzalez-Rodriguez, M. (2023). Empowering Education with Generative Artificial Intelligence Tools: Approach with an Instructional Design Matrix. *Sustainability*, *15*(15), 11524.

Sullivan, M., Kelly, A., & McLaughlan, P. (2023). ChatGPT in higher education: Considerations for academic integrity and student learning.

Vaswani, A., Shazeer, N., Parmar, N., Uszkoreit, J., Jones, L., Gomez, A. N., ... & Polosukhin, I. (2017). Attention is all you need. *Advances in neural information processing systems*, *30*.

Xames, M. D., & Shefa, J. (2023). ChatGPT for research and publication: Opportunities and challenges. *Available at SSRN 4381803*.

Zhai, X. (2022). ChatGPT user experience: Implications for education. *Available at SSRN 4312418*.

Zhao, W. X., Zhou, K., Li, J., Tang, T., Wang, X., Hou, Y., ... & Wen, J. R. (2023). A survey of large language models. *arXiv preprint arXiv:2303.18223*.